\documentclass{bmvc2k}

\usepackage{amsfonts}
\usepackage{multirow}
\usepackage{makecell}
\usepackage{booktabs}
\usepackage{graphicx}
\usepackage{amssymb}
\usepackage{tabularcalc}
\usepackage{placeins}
\usepackage{float}
\usepackage{xcolor}
\definecolor{ForestGreen}{RGB}{34,139,34}

\title{SiNeRF: Sinusoidal Neural Radiance Fields for Joint Pose Estimation and Scene Reconstruction}
\addauthor{Yitong Xia}{yitongxia@student.ethz.ch}{1}
\addauthor{Hao Tang}{hao.tang@vision.ee.ethz.ch}{1}
\addauthor{Radu Timofte}{radu.timofte@vision.ee.ethz.ch}{1, 2}
\addauthor{Luc Van Gool}{vangool@vision.ee.ethz.ch}{1}

\addinstitution{
    Computer Vision Laboratory\\
    D-ITET\\
    ETH Z\"urich\\Switzerland
}
\addinstitution{
    Computer Vision Laboratory\\
    CAIDAS, Institute of Computer Science\\
    University of W\"urzburg\\
    Germany
}

\runninghead{Y. Xia, H. Tang, R. Timofte, L. Van Gool}{SiNeRF}

\def\eg{\emph{e.g}\bmvaOneDot}

\def\etal{\emph{et al}\bmvaOneDot}

\begin{document}

\maketitle

\begin{abstract}
    NeRFmm \cite{wang2021nerf} is the Neural Radiance Fields (NeRF) that deal with \textit{Joint Optimization} tasks, i.e., reconstructing real-world scenes and registering camera parameters simultaneously. Despite NeRFmm producing precise scene synthesis and pose estimations, it still struggles to outperform the full-annotated baseline on challenging scenes. 
    In this work, we identify that there exists a systematic sub-optimality in joint optimization and further identify multiple potential sources for it. To diminish the impacts of potential sources, we propose \textit{Sinusoidal Neural Radiance Fields} (SiNeRF) that leverage sinusoidal activations for radiance mapping and a novel \textit{Mixed Region Sampling} (MRS) for selecting ray batch efficiently. 
    Quantitative and qualitative results show that compared to NeRFmm, SiNeRF achieves comprehensive significant improvements in image synthesis quality and pose estimation accuracy. Codes are available at \href{https://github.com/yitongx/sinerf}{https://github.com/yitongx/sinerf}.
\end{abstract}

\begin{figure}[htbp]
    \centering
    \resizebox{1\columnwidth}{!}{%
        \includegraphics[scale=1]{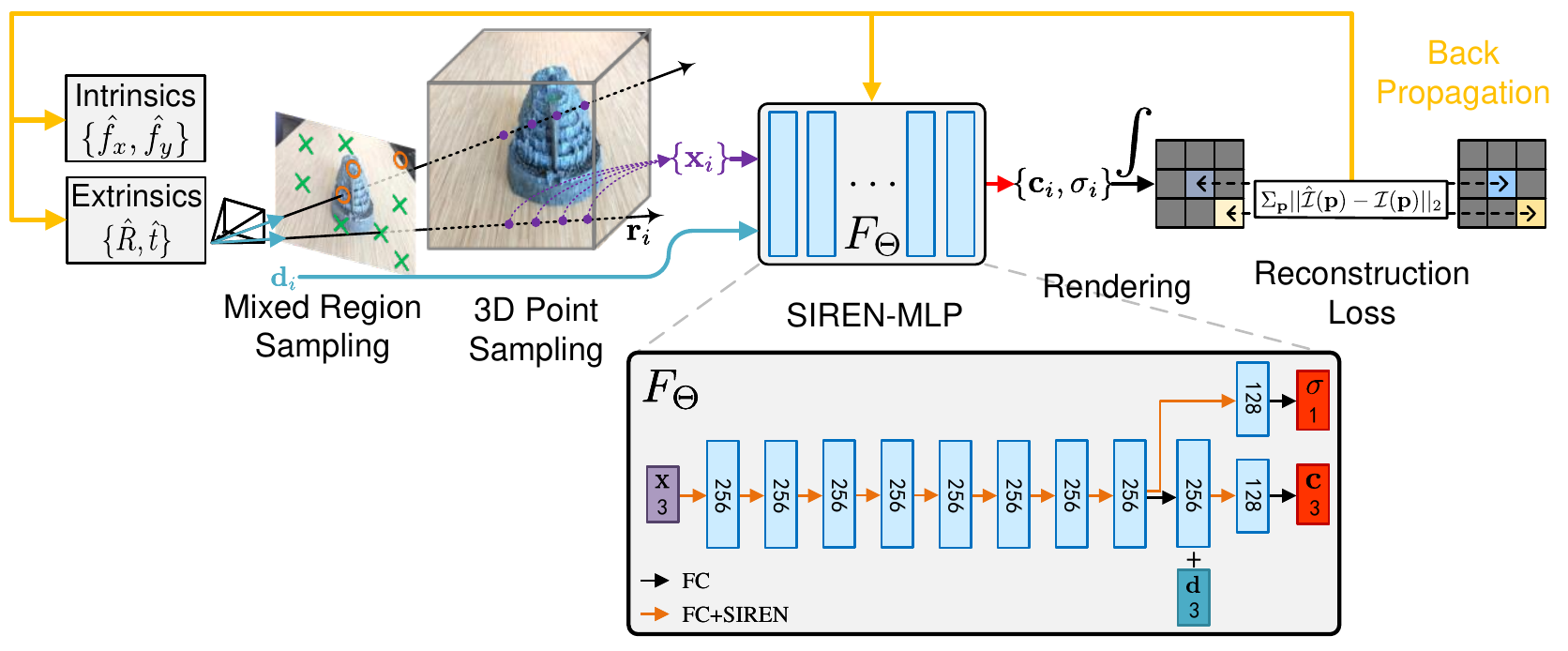}
    }
    \caption{
        General overview of SiNeRF. Our proposed \textit{Mixed Region Sampling} contains both key point ray candidates (in \color{orange} orange \color{black} circles) and random ray candidates (in \color{ForestGreen} green \color{black}crosses).
        The reconstruction loss updates both SiNeRF and camera parameters. We empirically scale $\sigma$ by 25 to avoid faded synthesis.
    }
    \vspace{-0.4cm}
    \label{fig:mlparch}
\end{figure}

\section{Introduction}
Adopting neural networks for Novel View Synthesis (NVS) has gained popularity. The community has achieved progress on various of representation forms, including multi-layer image plane \cite{srinivasan2019pushing, tucker2020single}, distance-based representation \cite{flynn2019deepview, shih20203d, riegler2021stable}, volume-based representation \cite{lombardi2019neural,mildenhall2020nerf}, etc. 
Among all, Neural Radiance Fields (NeRF) \cite{mildenhall2020nerf} are receiving growing attention for their concise structure and compelling synthesis image quality. NeRF implicitly represents scene space with a continuous radiance mapping function parameterized by a multi-layer perceptron (MLP), followed by volume rendering \cite{kajiya1984ray} to composite intermediate color and density outputs into final synthesis. 

Despite the compelling performances on scene reconstruction, NeRF-based methods are all trained on images with annotated camera parameters. Yet real-world scene images with precisely annotated camera parameters are always expensive and are not accessible all the time. 
NeRFmm \cite{wang2021nerf} proposes an end-to-end NVS framework without camera annotations, reconstructs high-fidelity real-world scenes, and estimates accurate poses comparable to a fully-annotated baseline. But NeRFmm is reported to struggle on textureless scenes where joint optimization can easily fall into local minima.

In this work, we aim to improve NeRFmm by alleviating its systematic sub-optimality. Inspired by the smooth nature and powerful expressiveness in complex signals of periodic activations, we design a novel \textbf{Si}nusoidal \textbf{Ne}ural \textbf{R}adiance \textbf{F}ields (SiNeRF) for joint optimization. 
We further reveal the inefficiency of conventional random sampling and propose a novel named \textit{Mixed Region Sampling} that allocates different weights to each pixel and samples from candidates strategically.

To conclude our contributions, in this work 
\begin{itemize}
    \item we propose a novel neural radiance field named \textit{SiNeRF} for alleviating the systematic sub-optimality of joint optimization in NeRFmm.
    \item we reveal the inefficiency of \textit{Random Sampling} and propose a novel \textit{Mixed Region Sampling} strategy that proved to be beneficial for tasks on challenging scenes. We prove that its combination with \textit{SiNeRF} provides the best performances.
    \item comprehensive quantitative, qualitative results, and ablation study on real-world scene dataset show our method's \textbf{comprehensive improvements} on camera pose estimation accuracy and novel view synthesis quality compared to NeRFmm.
\end{itemize}

\section{Related Work}
\noindent\textbf{Neural Scene Representation.}
Mildenhall \etal. \cite{mildenhall2020nerf} propose to encode scene representation inside a multi-layer perceptron (MLP) that directly regresses raw color and density. The final synthesis is composited by volume rendering \cite{kajiya1984ray, max1995optical} which is differentiable for backpropagating reconstruction loss. 
NeRF's success on high-fidelity NVS tasks has expanded its applications on series of vision tasks, \eg, scene relighting \cite{srinivasan2021nerv}, dynamic scene reconstruction \cite{ li2021neural, park2021nerfies,park2021hypernerf}, real-time scene synthesis \cite{hedman2021baking, muller2022instant}, etc.

\noindent\textbf{Scene Reconstruction with Imperfect Camera Annotations.}
Recently some NeRF-related works tackle scene reconstruction tasks without accurately annotated camera parameters. 
Our work is improved upon NeRFmm \cite{wang2021nerf}, which proposes an end-to-end framework that achieves compelling NVS performances without both camera intrinsics and extrinsics. 
A similar pipeline is proposed by iNeRF \cite{yen2021inerf}, yet it only estimates poses of unknown images, and its \textit{Interest Region Sampling} inspires us to improve sampling strategy for joint optimization. 
BARF \cite{lin2021barf} builds the connection between 2D image alignment and 3D scene reconstruction and uses a coarse-to-fine encoding adjustment for efficient training. Yet BARF is equipped with known intrinsic, and it initializes extrinsic with priors, whereas our work estimates both intrinsics and extrinsics with non-prior initializations.
SCNeRF \cite{jeong2021self} focuses on self-calibrating image distortions and does not output pose estimations, whose task concentrations are different from ours.

\noindent\textbf{Sinusoidal-Activated Multi-Layer Perceptron.}
SIREN \cite{sitzmann2020implicit} is the first to use sinusoidal-activated MLP for implicit neural representation. A SIREN-MLP is found to have rich expressiveness for representing zero- and first-order complex signals and thus relieving the network's hard prerequisite on Fourier-based input encoding. 
$\pi$-GAN \cite{chan2021pi} applies sinusoidal activations for Generative Adversarial Networks and achieves disentanglements on viewing-angle control and implicit 3D scene representation.
Inspired by prior works, to alleviate the sub-optimality of joint optimization in NeRFmm, we adopt sinusoidal activations into our SiNeRF for scene reconstruction and camera parameter estimations and further stabilize the training with our novel sampling strategy.

\section{Methods}
\label{sec:methods}
\subsection{NeRFmm Preliminary}
\label{sec:nerfmmpre}

NeRFmm \cite{wang2021nerf} reconstructs 3D scenes from sparse scene images ${\mathbf{I}} = \{ \mathcal{I}_1, \mathcal{I}_2, \dots, \mathcal{I}_N \}$ \textbf{without} annotated camera extrinsics $\mathbf{T} = \{ \mathcal{T}_1, \mathcal{T}_2, \dots, \mathcal{T}_N \}$ and intrinsics $\mathbf{f} = \{ f_x, f_y \}$. 
A continuous function, parameterized by a multilayer perceptron (MLP) $\Theta$, is used for view-dependent radiance mapping: $F_\Theta: (\mathbf{x}, \mathbf{d}) \to (\mathbf{c}, \sigma)$, 
where $\mathbf{x}$ is the point location in the implicit scene space, $\mathbf{d}$ is the corresponding unit-length viewing direction, $\mathbf{c}\in \mathbb{R}^3$ and $\sigma \in \mathbb{R}$ are the raw color and density values, respectively. 

The volume rendering \cite{kajiya1984ray, max1995optical}, denoted as operator $\mathcal{R}$, is used for compositing raw color and density values into final RGB pixels. Given a 2D pixel location $\mathbf{p} \in \mathbb{R}^2$ of the $i$-th image and a ray $\mathbf{r}(t) = \mathbf{o} + t\mathbf{d}$ that meets $\mathbf{p}$ on the image plane, the final color outputs would be:

\begin{equation}
    \mathcal{\hat{I}}_i(\mathbf{p}) = \mathcal{R}(\mathcal{T}_i, \mathbf{p}; \Theta) = \sum_{i=1}^{n} \, W_i \, (1 - \exp(- \sigma_i \delta_i)) \, \mathbf{c}_i, \;
    W_i = \exp(-\sum_{j=1}^{i-1}\, \sigma_j \, \delta_j),
\end{equation}
where $W_i$ is the accummulated transmittance of the ray $\mathbf{r}(t)$. $\mathbf{o} \in \mathbb{R}^3$ denotes the camera origin, $t \in [t_n, t_f]$ denotes the sampling point location within the nearest and farest field. 
$(\mathbf{c}_i, \sigma_i)$ are the raw color-density values of the $i$-th sampling point. $\delta_i=t_{i+1} - t_i$ is the interval between two sampling points along the ray. And we acquire a set of synthesized scene images $\hat{\mathbf{I}} = \{\hat{\mathcal{I}}_1, \hat{\mathcal{I}}_2, \dots, \hat{\mathcal{I}}_N \}$.

Now that the implicit mapping $F_\Theta$ and volume rendering operator $\mathcal{R}$ are differentiable, the pipeline is trained in a supervised learning fashion:
\begin{equation} \label{nerf_opt_obj}
    \Theta^{*}, \mathbf{T}^{*}, \mathbf{f}^{*} = \arg\min_{\Theta, \hat{\mathbf{T}}, \hat{\mathbf{f}}} L(\hat{\mathbf{I}}, \hat{\mathbf{T}}, \hat{\mathbf{f}} \,| \,\mathbf{I})
        = \arg\min_{\Theta, \hat{\mathbf{T}}, \hat{\mathbf{f}}} \sum_{i=1}^{N} \sum_{\mathbf{p}} \lVert \hat{\mathcal{I}}_i(\mathbf{p}) - \mathcal{I}_i(\mathbf{p}) \rVert_2^2.
\end{equation} 

\subsection{Camera Parameters Formation}
\label{sec:camera}

\noindent \textbf{Camera Intrinsics.}
Using a pinhole camera model, the camera intrinsics are represented in a calibration matrix:
\begin{equation}
    K = \left[\begin{matrix}
        \hat{f_x} & 0 & p_x \\
        0 & \hat{f_y} & p_y \\
        0 & 0 & 1 
    \end{matrix}
    \right],
\end{equation}
where $\hat{f_x}$ and $\hat{f_y}$ are optimizable focal length along horizontal and vertical axis respectively, $p_x$ and $p_y$ are the known coordinates of the image plane's origin. All intrinsics are shared across all input images.

\noindent \textbf{Camera Extrinsics.}
\label{sec:extrinsics}
Following prior works \cite{wang2021nerf,yen2021inerf}, the camera extrinsics in our work are represented by a rigid transform $T_{world} = [\mathbf{R}|\mathbf{t}]\in$ SE(3), where $\mathbf{R} \in$ SO(3) denotes the camera rotation and $\mathbf{t} \in \mathbb{R}^3$ denotes the camera translation. A vector $\mathbf{x}\in\mathbb{R}^3$ after rigid transform would be $\mathbf{x}' = \mathbf{R}\,\mathbf{x}+\mathbf{t}$.

To make the rotation parameters optimizable, by using Rodrigues' formula, we do axis-angle decomposition: 
\begin{equation} \label{rodrigues}
    \mathbf{R} \equiv e^{[\mathbf{r}]_{\times}} = \mathbf{I} + \frac{\sin\theta}{\theta}[\mathbf{r}]_{\times} + {\frac{1-\cos\theta}{\theta^2}}([\mathbf{r}]_{\times})^2,
\end{equation}
where $\mathbf{r}\in \mathbb{R}^3$,\, 
$\theta = \rVert \mathbf{r} \rVert$ denotes the angle of rotation, $\bar{\mathbf{r}}=\mathbf{r}/\rVert \mathbf{r} \rVert$ denotes the axis of rotation, and $[\cdot]_{\times}$ denotes the skew operator that converts a vector into a cross-product matrix. For each input image $\mathcal{I}_i$ we define its extrinsics $\{ \hat{\mathbf{r}}_i , \hat{\mathbf{t}}_i \}$, which can be directly optimized during training.

In our work we focus on jointly optimize $\{ \hat{f}_x, \hat{f}_y, \hat{\mathbf{r}}_i, \hat{\mathbf{t}}_i \}$ without using any two-stage refinements in NeRFmm \cite{wang2021nerf} or prior knowledge on intrinsics or extrinsics in BARF \cite{lin2021barf}.



\subsection{Improve Joint Optimization with SiNeRF}
\subsubsection{Potential Sub-Optimality of NeRFmm}
Although NeRFmm is capable of producing compelling camera parameter estimation and scene reconstruction, it suffers from falling into minima on specific scenes, \eg, textureless $Room$ scene and inconsistent $Orchids$ in the LLFF dataset. Moreover, in those scenes, there exist obvious gaps between COLMAP estimated intrinsics and estimated intrinsics as well as degeneration on NVS quality compared to ground truth, as reported in \cite{wang2021nerf}.

It has been a convention in NeRF-related tasks to use a 256-width ReLU-MLP for radiance mapping, while NeRFmm only adopts a 128-width ReLU-MLP for joint optimizing. Thus, it seems natural to blame the weak expressiveness of a narrow MLP for causing the potential sub-optimality. 
However, our experiment results show that simply increasing the width of MLP does not always improve NeRFmm's performances, \eg, there exists performance degenerations on $Fortress$ and $Orchids$ scenes, as shown in the \textit{ref-128} and \textit{ref-256} columns in Table \ref{tab_poses} and \ref{tab_nvs}.

To conclude, NeRFmm's joint optimization suffers from \textbf{a systematic sub-optimality} that cannot be solved by simply adopting a larger MLP. 

In the following sections, we identify that, the absence of a better radiance mapping network and the inefficient ray sampling techniques, would be two of the potential sources for such sub-optimality and they are the very focuses of our work.

\subsubsection{SiNeRF Architecture}
\label{sec:sirenmlp}
SiNeRF, our proposed radiance mapping network, consists of a SIREN-MLP \cite{sitzmann2020implicit} head followed by a color branch and a density branch.
We denote a $L$-layer SIREN-MLP head as:
\begin{equation}
    \Phi(\mathbf{x}) = \mathit{\phi}_{L} \circ \phi_{L-1} \circ \cdots \circ \phi_1(\mathbf{x}),
\end{equation}
where $\phi_l: \mathbb{R}^{d_{l-1}}\mapsto\mathbb{R}^{d_l}$ is the $l$-th fully-connected SIREN layer
\begin{equation}
    \phi_l(\mathbf{x})
    = \sin(\alpha_l \, (\mathbf{W}_l \, \mathbf{x} + \mathbf{b}_l) + \beta_l),
\end{equation}
defined by a weight $\mathbf{W}_l \in \mathbb{R}^{d_l\times d_{l-1}}$, a bias $\mathbf{b}_l \in \mathbb{R}^{d_l}$, a frequency scaling factor $\alpha_l \in \mathbb{R}$, and a phase shift factor $\beta_l \in \mathbb{R}^{d_l}$.

We follow the initialization scheme of \cite{sitzmann2020implicit}. We set frequency scaling factors $\alpha_1 = 30$ and $\alpha_l = 1$ for $l\in\{2, 3, \cdots, L\}$. We set all phase shift factors $\beta_l = \mathbf{0}$ for $l \in \{1, \cdots, L\}$. 
We initialize weight matrices $\mathbf{W}_1 \sim \mathcal{U}(-1/d_1, 1/d_1)$ and $\mathbf{W}_l \sim \mathcal{U}(-\sqrt{6/d_l}, \sqrt{6/d_l})$ for $l\in\{2, 3, \cdots, L\}$. For SiNeRF in experiments, we keep layer number $L=8$ and hidden unit number $d_l=256$ for $l\in\{1, 2, \cdots, L\}$. No positional encoding is used for input.

After the SIREN-MLP head, we append two branches for outputting raw color $\mathbf{c}\in\mathbb{R}^3$ and raw density $\sigma\in\mathbb{R}$ respectively. Please see Figure\,\ref{fig:mlparch} for more details about SiNeRF architecture.

\subsubsection{Partial Benefits For the Joint Optimization}
We observe that simply replacing ReLU-MLP with SiNeRF for radiance mapping can \textbf{only} improve performances on a limited number of scenes. The reason for the \textbf{partially} improvements can be that: the input signal is clamped by sinusoidal activation’s amplitude and limiting the output value range may bring stability for optimization while also may reduce the signal’s expressiveness. 
Thus, to achieve the general alleviation of sub-optimality in joint optimization, we propose improvements on the ray sampling strategy in the following section.

\subsection{Mixed Region Sampling (MRS)}
\label{sec:mrs}

Random Sampling is to randomly choose $M$ rays from random candidate set $\mathcal{P}_{random}^{(i)} = \{\mathbf{p}\,|\,\forall \mathbf{p} \in \mathcal{I}_i\}$. Along each ray, we then apply 3D point sampling to select several spatial points for radiance mapping. 
Recall that NeRF's reconstruction loss computes the average pixel difference between rendered pixels and corresponding ground truth. A batch of randomly sampled rays would have different colors. \textbf{It guarantees the ray batch's diversity and is a critical condition for efficient training.}

To prove that, we set a simple experiment NeRFmm on \textit{Flower} scene and fix all settings unchanged, except that every ray batch is selected from a randomly-placed $32\times32$ image patch instead of randomly-selected 1024 rays. Such sampling strategy is the most extreme case where the ray batch has minimum diversity. As for results, the PSNR score halves and pose errors increase by 20 times, compared to baseline.

Random Sampling is straightforward and has been widely adopted by NeRFmm and other NeRF-related works \cite{lin2021barf,wang2021nerf,mildenhall2020nerf}. Yet this strategy still has its limitation.
On a textureless scene like $Fortress$ in the LLFF dataset, even a batch of randomly sampled pixels may have homogenized colors, which will result in an iteration of "\textit{poor supervision}", i.e., the network is not forced to produce discriminative outputs at different pixel locations.


We argue that, on the one hand, there are fewer constraints for joint optimization tasks. The optimizing direction provided by such poor supervision might potentially make models easier to fall into local minima. 
On the other hand, inspired by offline SfM methods \cite{triggs1999bundle,schoenberger2016sfm} that leverage key points for efficient inter-image matching, we believe that, compared to Random Sampling with equal weights, treating pixels with different importances and sampling them strategically would help alleviating the sub-optimality in joint optimization.

Thus, we propose a novel sampling strategy designed for joint optimization tasks, named \textit{Mixed Region Sampling} (MRS):

\begin{itemize}
    \item For the $i$-th image, we use a SIFT detector to find a set of keypoints $\mathcal{P}_0^{(i)} = \{\bar{\mathbf{p}}_1, \bar{\mathbf{p}}_2, \cdots, \bar{\mathbf{p}}_K\}$. 
    For each keypoint $\bar{\mathbf{p}}_j$ we form a local region set $\mathcal{P}_j^{(i)} = \{\mathbf{p}\,|\,\forall \mathbf{p} \in \overline{\mathcal{N}}_{5\times 5}(j) \}$ where $\overline{\mathcal{N}}_{5\times 5}(j)$ denotes a set of neighbour points of keypoint $\bar{\mathbf{p}}_j$ within a $5 \times 5$ region. The region candidate set is defined as $\mathcal{P}_{region}^{(i)} = \mathcal{P}_0^{(i)} \cup \mathcal{P}_1^{(i)} \cup \cdots \cup \mathcal{P}_K^{(i)}$. 
    \item Then, we define a time-variant weight for region sampling:
    \begin{equation}
        w(t) = 
        \begin{cases}
            1 - t/t_r & 0 \leq t \leq t_r\\
            0 &  t > t_r
        \end{cases},
    \end{equation}
    that linearly decreases within the range $[0, t_r]$, where $t_r$ represents the end time step of MRS. For the $M$-ray sampling at time $t$, $w(t)M$ rays are sampled from region set $\mathcal{P}_{region}^{(i)}$ and $(1-w(t))M$ rays are sampled from random set $\mathcal{P}_{random}^{(i)}$. After $t_r$, MRS falls back to random sampling.
    
\end{itemize}

Our \textit{Mixed Region Sampling} (MRS) leverages both region candidates for efficient region matching at the early training stage and random candidates for discriminative learning at the late training stage, which is improved upon the \textit{Interst Region Sampling} \cite{yen2021inerf} that only samples candidates from the regions of interest for pose optimization.
We show in the ablation study in Section\,\ref{sec:ablation} that adopting MRS is critical for improving both image quality and pose accuracy, and its combination with SIREN-MLP provides more general alleviation on sub-optimality in joint optimization.

\subsection{Testing}
Because of the ambiguity between camera translation scale and camera intrinsics \cite{pollefeys1997stratified}, the learned poses may not be in the same pose space with COLMAP annotated poses. Thus, \textit{pose alignment} is required for a valid test. 
Following \cite{wang2021nerf}, firstly, we use ATE toolbox \cite{zhang2018tutorial} to compute a Sim(3) transformation that aligns the COLMAP testing trajectory with the testing trajectory in the SiNeRF pose space. 
Secondly, we optimize the roughly-aligned testing trajectories by minimizing reconstruction loss while keeping intrinsics and network parameters fixed. This step is to provide precise alignments on trajectories for testing.
Lastly, we compute the image quality and pose metrics on test images.

\section{Experiments}
\subsection{Settings}
\noindent\textbf{Dataset.}
We experiment on the LLFF dataset with 8 forward-facing real-world scenes. Every 8-th image in the image sequences is selected for testing. All image resolution is set to $756\times 1008$. Camera annoations are estimated by COLMAP \cite{schoenberger2016sfm}.

\noindent\textbf{Training Details.}
For intrinsics, we initialize $\hat{f}_x$ and $\hat{f}_y$ to be the width and height of the image. 
For extrinsics, for each image $\mathcal{I}_i$ we initialize the translation $\hat{\mathbf{t}}_i$ to be a zero vector, and rotation matrix $\mathbf{R}_i$ to be an identity matrix, i.e., $\hat{\mathbf{r}}_i$ to be a zero vector. We initialize an 8-layer 256-width SIREN-MLP with methods mentioned in Section\,\ref{sec:sirenmlp}. 
For each iteration, 1024 rays are selected by \textit{Mixed Region Sampling} (MRS), where $t_r$ is set to $500$ epochs for $Fortress$ and $Trex$ scenes and $50$ epochs for the rest. Along each ray 128 coordinates are uniformly selected \textbf{without} using hierarchical sampling \cite{mildenhall2020nerf}. 

We train the model for 10k epochs for each scene with three Adam \cite{kingma2014adam} optimizers for intrinsics, extrinsics, and SIREN-MLP, respectively.
Extrinsics' and intrinsics' learning rates are initialized to 1e-3 and exponentially decay by 0.9 every 100 epochs. 
SIREN-MLP's learning rate is initialized to 1e-3 for all scenes except that we lower it to 5e-4 for $Fortress$ and 1e-4 for $Orchids$ scene, and exponentially decay by 0.9954 every 10 epochs.

\vspace{-0.2cm}
\subsection{Quantitative Evaluations}
\label{quantitative}
We compare the performances between NeRFmm baselines and SiNeRF. Mean pose translation and rotation errors are shown in Table\,\ref{tab_poses}. NVS image qualities on three metrics PSNR, SSIM \cite{wang2004image_ssim} and LPIPS \cite{zhang2018unreasonable_lpips} are shown in Table\,\ref{tab_nvs}. 

As shown in the results, adopting a wider network does not necessarily improve the pose accuracy (\eg, $Fern$ scene) and image quality (\eg, $Fortress$ scene), indicating that joint optimization may exist a systematic sub-optimality. Meanwhile, SiNeRF improves image qualities significantly while achieving pose estimations closed to the ground truth provided by COLMAP.

We mention that the pose errors only indicate how well our pose estimations match the COLMAP estimations. Small pose errors do not guarantee good NVS image qualities. 

Our method does not outperform NeRFmm256 baseline on \textit{Leaves} scenes, which indicates that joint optimization is sensitive to scene content. The systematic sub-optimality can only be alleviated instead of completely solved by SiNeRF.

\begin{table}[]
    \centering
    \resizebox{1\columnwidth}{!}{%
        \begin{tabular}{ccccccc}
            \Xhline{2\arrayrulewidth}
            \multicolumn{1}{c}{\multirow{3}{*}{\textbf{Scene}}} & \multicolumn{6}{c}{\textbf{Pose Error}} \\
            \cmidrule(lr){2-7}
            \multicolumn{1}{c}{} & \multicolumn{3}{c}{Translation($\times 10^{-2}$) $\downarrow$} & \multicolumn{3}{c}{Rotation($^\circ$) $\downarrow$} \\
            \cmidrule(lr){2-4} \cmidrule(lr){5-7}

            \multicolumn{1}{c}{} & \textit{NeRFmm128} & \textit{NeRFmm256} & \textbf{SiNeRF} & \textit{NeRFmm128} & \textit{NeRFmm256} & \textbf{SiNeRF} \\
            \Xhline{2\arrayrulewidth}
            Fern & 0.514 & 0.765 & \textbf{0.438} & 0.957 & 1.566 & \textbf{0.743} \\
            Flower & 1.039 & 1.200 & \textbf{0.796} & 3.657 & 3.211 & \textbf{0.506} \\
            Fortress & 6.463 & 6.046 & \textbf{4.068} & 2.590 & 2.410 & \textbf{1.772} \\
            Horns & 1.607 & \textbf{1.476} & 2.153 & 3.806 & 3.044 & \textbf{2.662} \\
            Leaves & 0.676 & \textbf{0.608} & 0.831 & 8.248 & \textbf{6.782} & 8.762 \\
            Orchids & 1.627 & 2.243 & \textbf{1.257} & 4.140 & 5.459 & \textbf{3.244} \\
            Room & \textbf{1.315} & 2.148 & 2.145 & 3.357 & 3.745 & \textbf{2.075} \\
            Trex & 1.213 & 1.467 & \textbf{0.462} & 4.953 & 6.339 & \textbf{0.856} \\
            \hline
            Mean & 1.807 & 1.994 & \textbf{1.519} & 3.964 & 4.070 & \textbf{2.578} \\
            \Xhline{2\arrayrulewidth}
            \vspace{0.001cm}
        \end{tabular}
    }
    \caption{Quantitative results of pose estimation on LLFF dataset. \textit{NeRFmm128} and \textit{NeRFmm256} denote the NeRFmm baseline with MLP width to be 128 and 256 respectively. Best results are \textbf{bolded}.}
    \label{tab_poses}
\end{table}

\begin{table}[]
    \resizebox{1\columnwidth}{!}{%
        \centering
        \begin{tabular}{cccccccccc}
            \Xhline{2\arrayrulewidth}
            \multicolumn{1}{c}{\multirow{3}{*}{\textbf{Scene}}} & \multicolumn{9}{c}{\textbf{Image Quality}} \\
            \cmidrule(lr){2-10}
            \multicolumn{1}{c}{} & \multicolumn{3}{c}{PSNR $\uparrow$} & \multicolumn{3}{c}{SSIM $\uparrow$} & \multicolumn{3}{c}{LPIPS $\downarrow$} \\
            \cmidrule(lr){2-4} \cmidrule(lr){5-7} \cmidrule(lr){8-10}

            \multicolumn{1}{c}{} & \textit{NeRFmm128} & \textit{NeRFmm256} & \textbf{SiNeRF} & \textit{NeRFmm128} & \textit{NeRFmm256} & \textbf{SiNeRF} & \textit{NeRFmm128} & \textit{NeRFmm256} & \textbf{SiNeRF} \\
            \Xhline{2\arrayrulewidth}
            Fern & 21.811 & 22.154 & \textbf{22.482} & 0.631 & 0.648 & \textbf{0.665} & 0.479 & 0.459 & \textbf{0.437} \\
            Flower & 25.430 & 26.606 & \textbf{27.229} & 0.714 & 0.772 & \textbf{0.798} & 0.366 & 0.296 & \textbf{0.295} \\
            Fortress & 26.173 & 25.596 & \textbf{27.465} & 0.653 & 0.602 & \textbf{0.722} & 0.438 & 0.538 & \textbf{0.393} \\
            Horns & 22.949 & 23.174 & \textbf{24.142} & 0.626 & 0.635 & \textbf{0.684} & 0.492 & 0.506 & \textbf{0.431} \\
            Leaves & 18.647 & \textbf{19.741} & 19.152 & 0.512 & \textbf{0.609} & 0.571 & 0.476 & \textbf{0.385} & 0.392 \\
            Orchids & 16.695 & 15.858 & \textbf{16.922} & 0.391 & 0.350 & \textbf{0.408} & 0.540 & 0.550 & \textbf{0.529} \\
            Room & 25.623 & 25.675 & \textbf{26.101} & 0.831 & 0.836 & \textbf{0.844} & 0.450 & \textbf{0.411} & 0.426 \\
            Trex & 22.551 & 23.376 & \textbf{24.939} & 0.719 & 0.759 & \textbf{0.816} & 0.438 & 0.390 & \textbf{0.356} \\
            \hline
            Mean & 22.485 & 22.773 & \textbf{23.554} & 0.635 & 0.651 & \textbf{0.689} & 0.460 & 0.442 & \textbf{0.407} \\
            \Xhline{2\arrayrulewidth}
        \end{tabular}
    }
    \caption{Quantitative results of novel view synthesis on LLFF dataset.
    \textit{NeRFmm128} and \textit{NeRFmm256} denote the NeRFmm baseline with MLP width to be 128 and 256 respectively. 
    Best results are \textbf{bolded}.}
    \label{tab_nvs}
\end{table}

\subsection{Qualitative Results}
\begin{figure}[]
    \centering
    \begin{minipage}{.02\textwidth}
        \centering
        \rotatebox{90}{Fern}
    \end{minipage}
    \begin{minipage}{.2\textwidth}
        \centering
        \includegraphics[width=\textwidth]{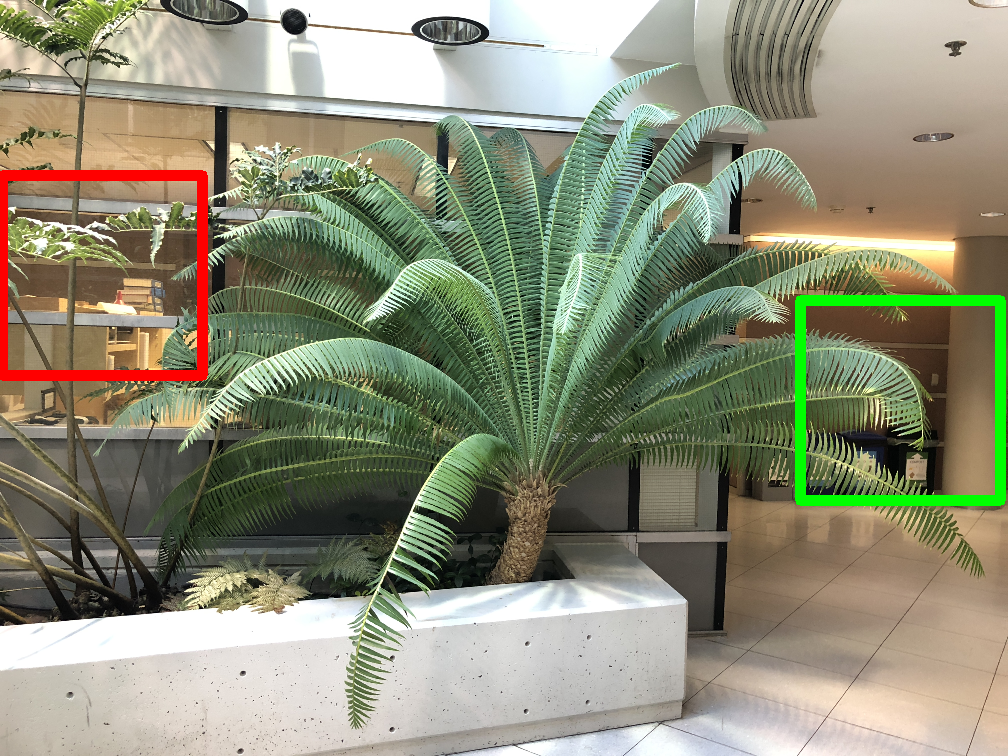}
    \end{minipage}    
    \begin{minipage}{.15\textwidth}
        \centering
        \includegraphics[width=\textwidth]{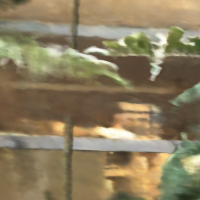}
    \end{minipage}
    \begin{minipage}{.15\textwidth}
        \centering
        \includegraphics[width=\textwidth]{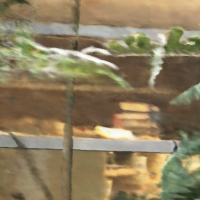}
    \end{minipage}
    \begin{minipage}{.15\textwidth}
        \centering
        \includegraphics[width=\textwidth]{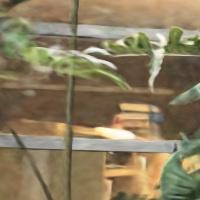}
    \end{minipage}
    \begin{minipage}{.15\textwidth}
        \centering
        \includegraphics[width=\textwidth]{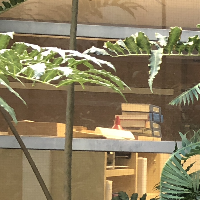}
    \end{minipage} \\
    \vspace{0.05cm}

    \begin{minipage}{.02\textwidth}
        \centering
        \rotatebox{90}{$ $}
    \end{minipage}
    \begin{minipage}{.2\textwidth}
        \centering
        \includegraphics[width=\textwidth]{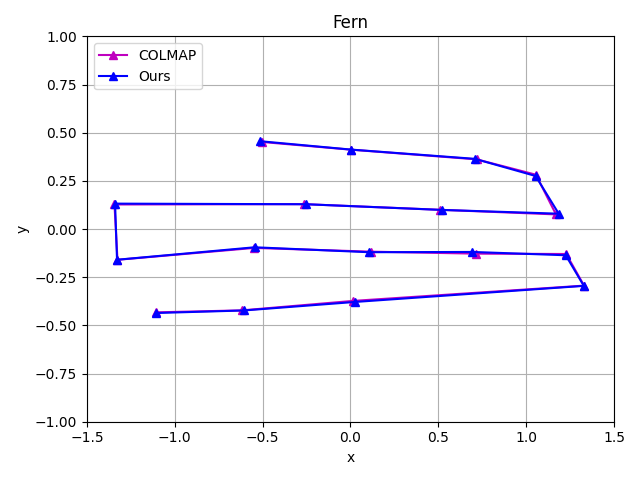}
    \end{minipage}    
    \begin{minipage}{.15\textwidth}
        \centering
        \includegraphics[width=\textwidth]{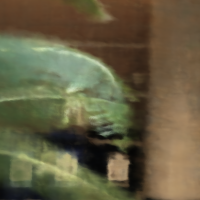}
    \end{minipage}
    \begin{minipage}{.15\textwidth}
        \centering
        \includegraphics[width=\textwidth]{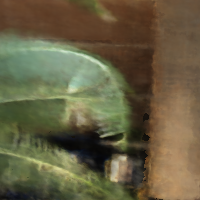}
    \end{minipage}
    \begin{minipage}{.15\textwidth}
        \centering
        \includegraphics[width=\textwidth]{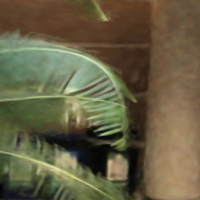}
    \end{minipage}
    \begin{minipage}{.15\textwidth}
        \centering
        \includegraphics[width=\textwidth]{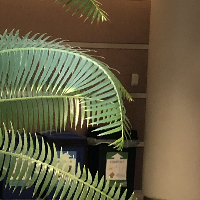}
    \end{minipage}

    \vspace{0.2cm}

    \begin{minipage}{.02\textwidth}
        \centering
        \rotatebox{90}{Flower}
    \end{minipage}
    \begin{minipage}{.2\textwidth}
        \centering
        \includegraphics[width=\textwidth]{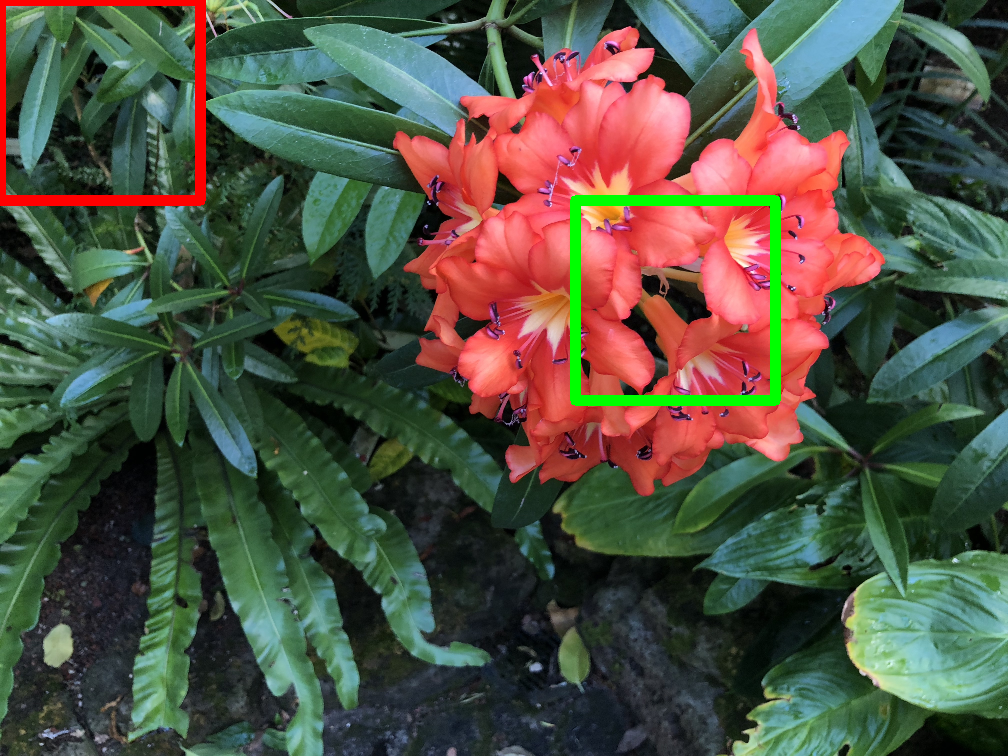}
    \end{minipage}    
    \begin{minipage}{.15\textwidth}
        \centering
        \includegraphics[width=\textwidth]{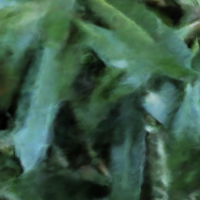}
    \end{minipage}
    \begin{minipage}{.15\textwidth}
        \centering
        \includegraphics[width=\textwidth]{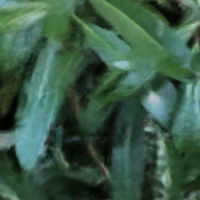}
    \end{minipage}
    \begin{minipage}{.15\textwidth}
        \centering
        \includegraphics[width=\textwidth]{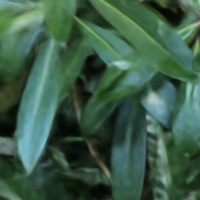}
    \end{minipage}
    \begin{minipage}{.15\textwidth}
        \centering
        \includegraphics[width=\textwidth]{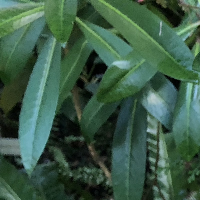}
    \end{minipage} \\
    \vspace{0.05cm}

    \begin{minipage}{.02\textwidth}
        \centering
        \rotatebox{90}{$ $}
    \end{minipage}
    \begin{minipage}{.2\textwidth}
        \centering
        \includegraphics[width=\textwidth]{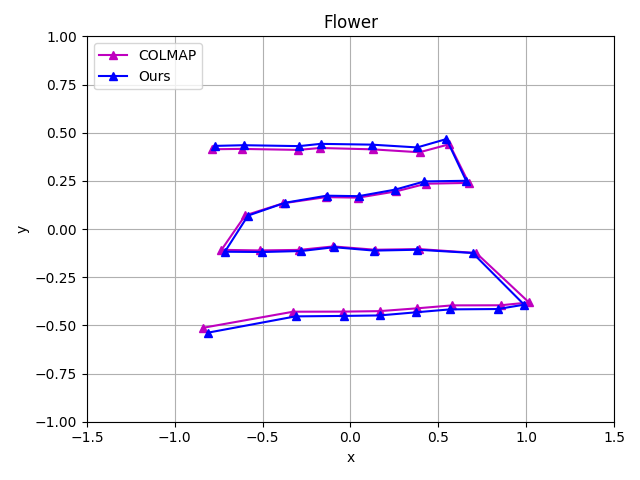}
    \end{minipage}    
    \begin{minipage}{.15\textwidth}
        \centering
        \includegraphics[width=\textwidth]{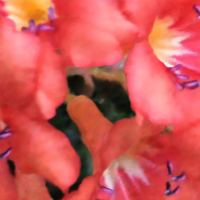}
    \end{minipage}
    \begin{minipage}{.15\textwidth}
        \centering
        \includegraphics[width=\textwidth]{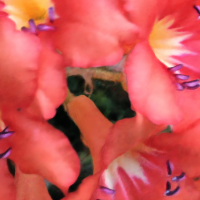}
    \end{minipage}
    \begin{minipage}{.15\textwidth}
        \centering
        \includegraphics[width=\textwidth]{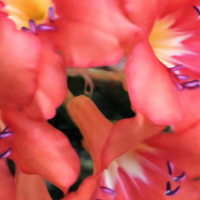}
    \end{minipage}
    \begin{minipage}{.15\textwidth}
        \centering
        \includegraphics[width=\textwidth]{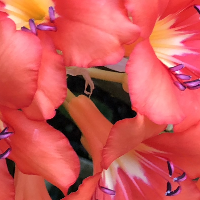}
    \end{minipage}

    \vspace{0.2cm}

    \begin{minipage}{.02\textwidth}
        \centering
        \rotatebox{90}{Horns}
    \end{minipage}
    \begin{minipage}{.2\textwidth}
        \centering
        \includegraphics[width=\textwidth]{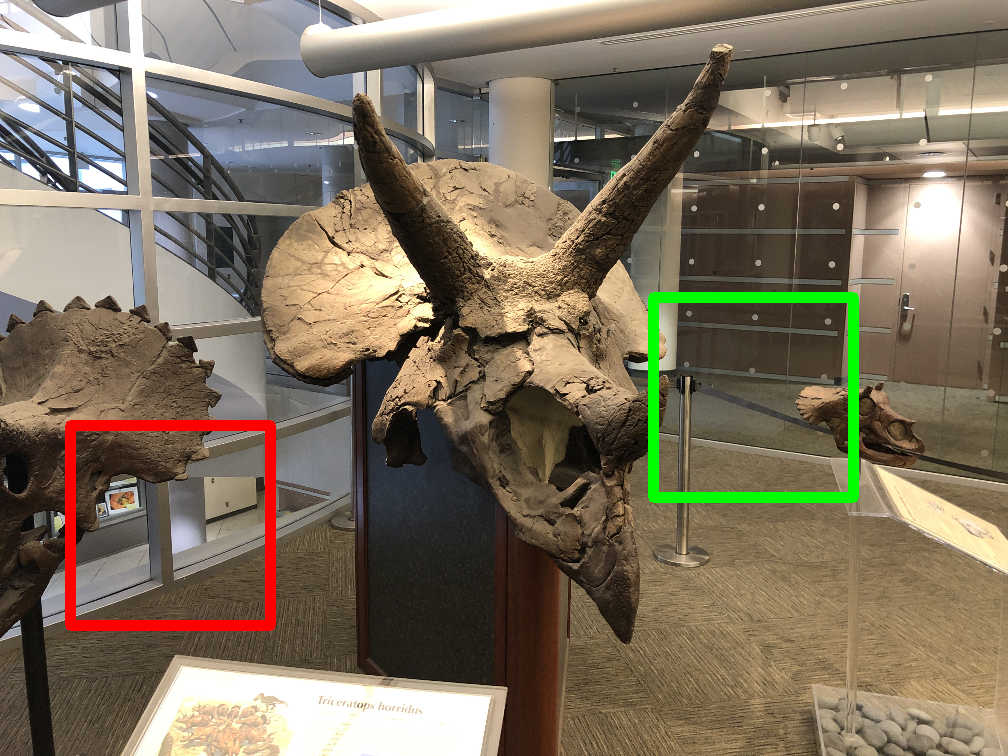}
    \end{minipage}    
    \begin{minipage}{.15\textwidth}
        \centering
        \includegraphics[width=\textwidth]{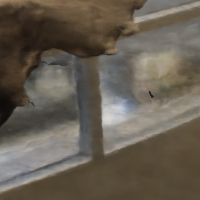}
    \end{minipage}
    \begin{minipage}{.15\textwidth}
        \centering
        \includegraphics[width=\textwidth]{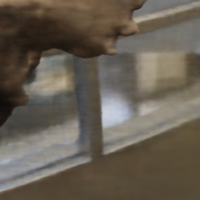}
    \end{minipage}
    \begin{minipage}{.15\textwidth}
        \centering
        \includegraphics[width=\textwidth]{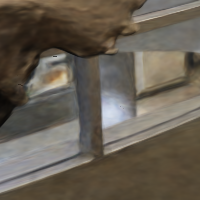}
    \end{minipage}
    \begin{minipage}{.15\textwidth}
        \centering
        \includegraphics[width=\textwidth]{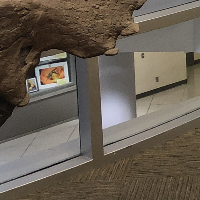}
    \end{minipage} \\
    \vspace{0.05cm}

    \begin{minipage}{.02\textwidth}
        \centering
        \rotatebox{90}{$ $}
    \end{minipage}
    \begin{minipage}{.2\textwidth}
        \centering
        \includegraphics[width=\textwidth]{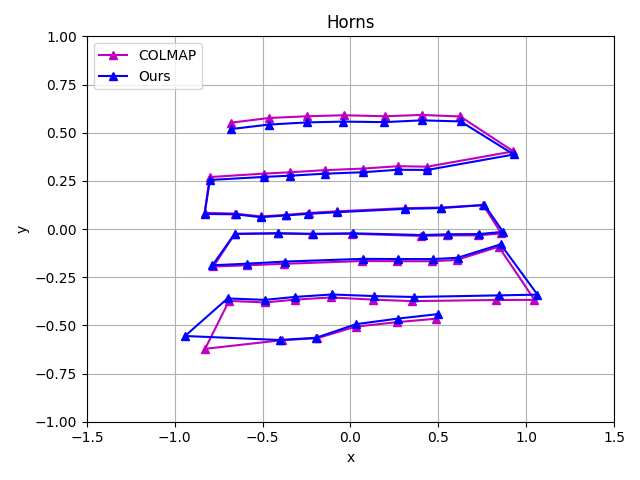}
    \end{minipage}    
    \begin{minipage}{.15\textwidth}
        \centering
        \includegraphics[width=\textwidth]{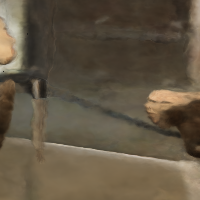}
    \end{minipage}
    \begin{minipage}{.15\textwidth}
        \centering
        \includegraphics[width=\textwidth]{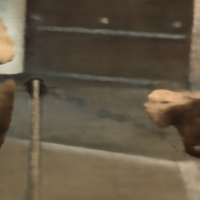}
    \end{minipage}
    \begin{minipage}{.15\textwidth}
        \centering
        \includegraphics[width=\textwidth]{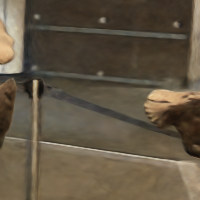}
    \end{minipage}
    \begin{minipage}{.15\textwidth}
        \centering
        \includegraphics[width=\textwidth]{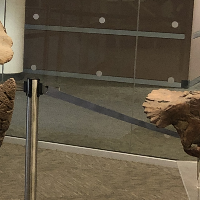}
    \end{minipage}

    \vspace{0.2cm}
    \begin{minipage}{.02\textwidth}
        \centering
        \rotatebox{90}{Trex}
    \end{minipage}
    \begin{minipage}{.2\textwidth}
        \centering
        \includegraphics[width=\textwidth]{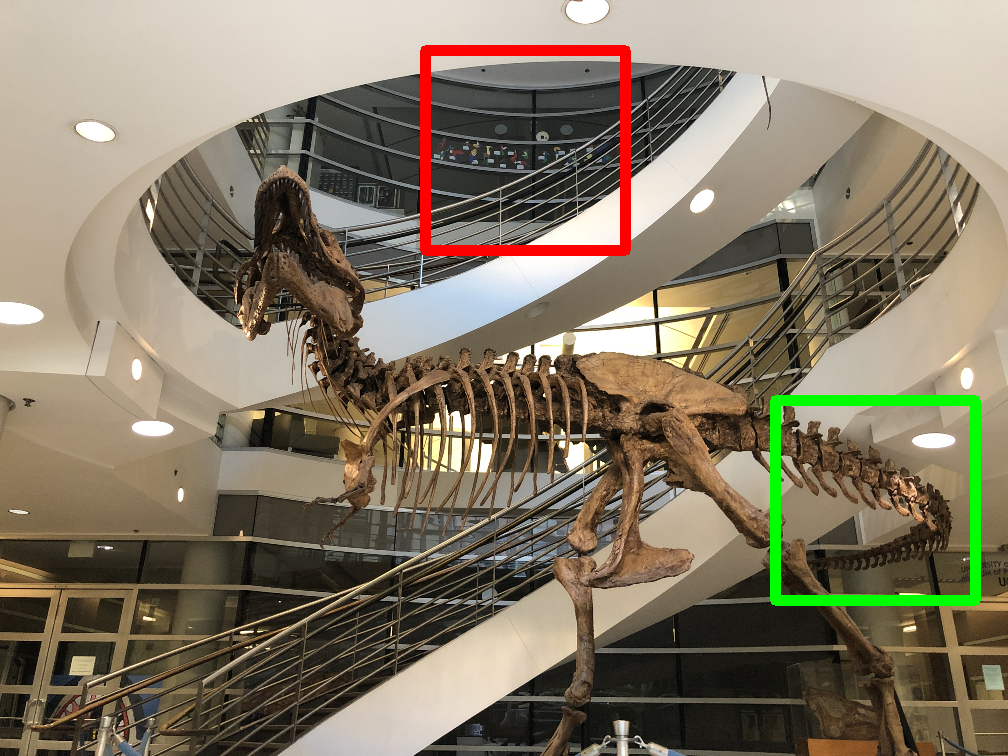}
    \end{minipage}    
    \begin{minipage}{.15\textwidth}
        \centering
        \includegraphics[width=\textwidth]{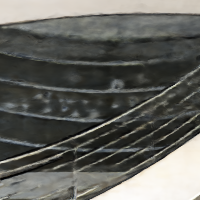}
    \end{minipage}
    \begin{minipage}{.15\textwidth}
        \centering
        \includegraphics[width=\textwidth]{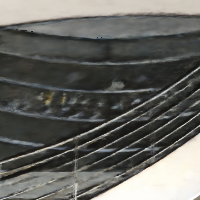}
    \end{minipage}
    \begin{minipage}{.15\textwidth}
        \centering
        \includegraphics[width=\textwidth]{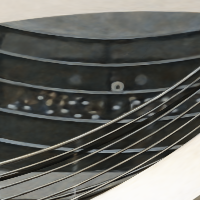}
    \end{minipage}
    \begin{minipage}{.15\textwidth}
        \centering
        \includegraphics[width=\textwidth]{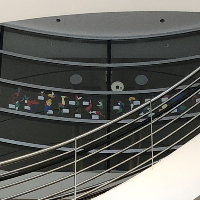}
    \end{minipage} \\
    \vspace{0.05cm}

    \begin{minipage}{.02\textwidth}
        \centering
        \rotatebox{90}{$ $}
    \end{minipage}
    \begin{minipage}{.2\textwidth}
        \centering
        \includegraphics[width=\textwidth]{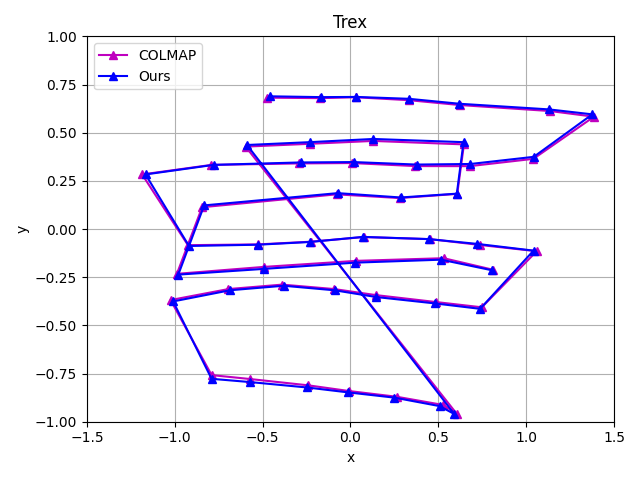}
    \end{minipage}    
    \begin{minipage}{.15\textwidth}
        \centering
        \includegraphics[width=\textwidth]{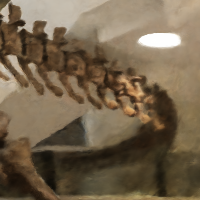}
    \end{minipage}
    \begin{minipage}{.15\textwidth}
        \centering
        \includegraphics[width=\textwidth]{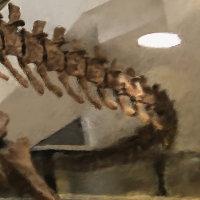}
    \end{minipage}
    \begin{minipage}{.15\textwidth}
        \centering
        \includegraphics[width=\textwidth]{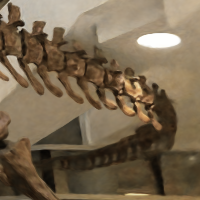}
    \end{minipage}
    \begin{minipage}{.15\textwidth}
        \centering
        \includegraphics[width=\textwidth]{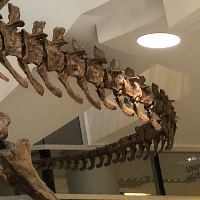}
    \end{minipage}

    \vspace{0.2cm}

    \begin{minipage}{.02\textwidth}
        \centering
        \rotatebox{90}{$ $}
    \end{minipage}
    \begin{minipage}{.2\textwidth}
        \centering
        \small{$ $}
    \end{minipage}    
    \begin{minipage}{.15\textwidth}
        \centering
        \small{NeRFmm-128}
    \end{minipage}
    \begin{minipage}{.15\textwidth}
        \centering
        \small{NeRFmm-256}
    \end{minipage}
    \begin{minipage}{.15\textwidth}
        \centering
        \small{\textbf{SiNeRF}}
    \end{minipage}
    \begin{minipage}{.15\textwidth}
        \centering
        \small{Reference}
    \end{minipage} 

    \caption{Qualitative results of our method on the LLFF dataset. Comparisons on pose trajectories between SiNeRF and COLMAP are displayed in the bottom-left corner for each scene.}
    \label{fig:qualitative}
\end{figure}

In Figure\,\ref{fig:qualitative} we display the comparisons on image synthesis between NeRFmm baselines and SiNeRF. Our method is able to reconstruct fine details in the real-world scene with high fidelity.

We also display the comparisons on pose trajectories between SiNeRF estimations and COLMAP estimations. The highly overlapped trajectories show that our method can learn accurate pose estimations closed to classical SfM estimations.

\begin{table}[]
    \centering
    \resizebox{1\columnwidth}{!}{%
        \centering
        \begin{tabular}{cccccccc}
            \Xhline{2\arrayrulewidth}
            \multicolumn{1}{c}{\multirow{2}{*}{\textbf{Scene}}} & \multicolumn{1}{c}{\multirow{2}{*}{\textbf{Items}}} & \multicolumn{2}{c}{\textbf{Pose Error}} & \multicolumn{3}{c}{\textbf{Image Quality}} \\
            \cmidrule(lr){3-4} \cmidrule(lr){5-7}
            \multicolumn{1}{c}{} & \multicolumn{1}{c}{} & Translation($\times 10^{-2}$) $\downarrow$ & Rotation($^\circ$) $\downarrow$ & \;\;PSNR $\uparrow$\;\; & \;\;SSIM $\uparrow$\;\; & \;\;LPIPS $\downarrow$\;\; \\
            \Xhline{2\arrayrulewidth}
            \multirow{4}{*}{Fortress} & \textbf{SiNeRF} & \textbf{4.068} & \textbf{1.772} & \textbf{27.465} & \textbf{0.722} & \textbf{0.393} \\
            {} & \textit{w/o SIREN} & 18.005 & 155.553 & 18.593 & 0.492 & 0.484 \\
            {} & \textit{w/o MRS} & 6.242 & 1.797 & 25.542 & 0.605 & 0.532 \\
            {} & \textit{w/o SIREN and MRS} & 6.046 & 2.410 & 25.596 & 0.602 & 0.538  \\
            \hline
            \multirow{4}{*}{Trex} & \textbf{SiNeRF} & \textbf{0.462} & \textbf{0.856} & \textbf{24.939} & \textbf{0.816} & \textbf{0.356} \\
            {} & \textit{w/o SIREN} & 1.891 & 7.958 & 22.478 & 0.719 & 0.430 \\
            {} & \textit{w/o MRS} & 17.755 & 133.462 & 14.984 & 0.452 & 0.659 \\
            {} & \textit{w/o SIREN and MRS} & 1.467 & 6.339 & 23.376 & 0.759 & 0.390 \\
            \Xhline{2\arrayrulewidth} 
            \vspace{0.001cm} 
        \end{tabular}
    }
    \caption{Quantitative results of ablation study. (1) \textit{w/o SIREN} denotes NeRFmm with 256-width ReLU-MLP and MRS. (2) \textit{w/o MRS} denotes SIREN-MLP with Random Sampling. (3) \textit{w/o SIREN and MRS} denotes NeRFmm with 256-with ReLU-MLP and Random Sampling, which is the baseline. For all MRS in the table we set $t_r=500$. Best results are \textbf{bolded}.}
    \label{tab_ablation}
\end{table}

\subsection{Ablation Study}
\label{sec:ablation}
To exclude the influence of adopting a wider MLP, we list 256-width ReLU-MLP results in the \textit{NeRFmm256} columns in Table\,\ref{tab_poses} and \ref{tab_nvs}.

\begin{figure}[]
    \centering
    \begin{minipage}{.025\textwidth}
        \rotatebox{90}{Image}
    \end{minipage}%
    \begin{minipage}{.22\textwidth}
        \centering
        \small \textbf{SiNeRF}
        \includegraphics[width=\textwidth]{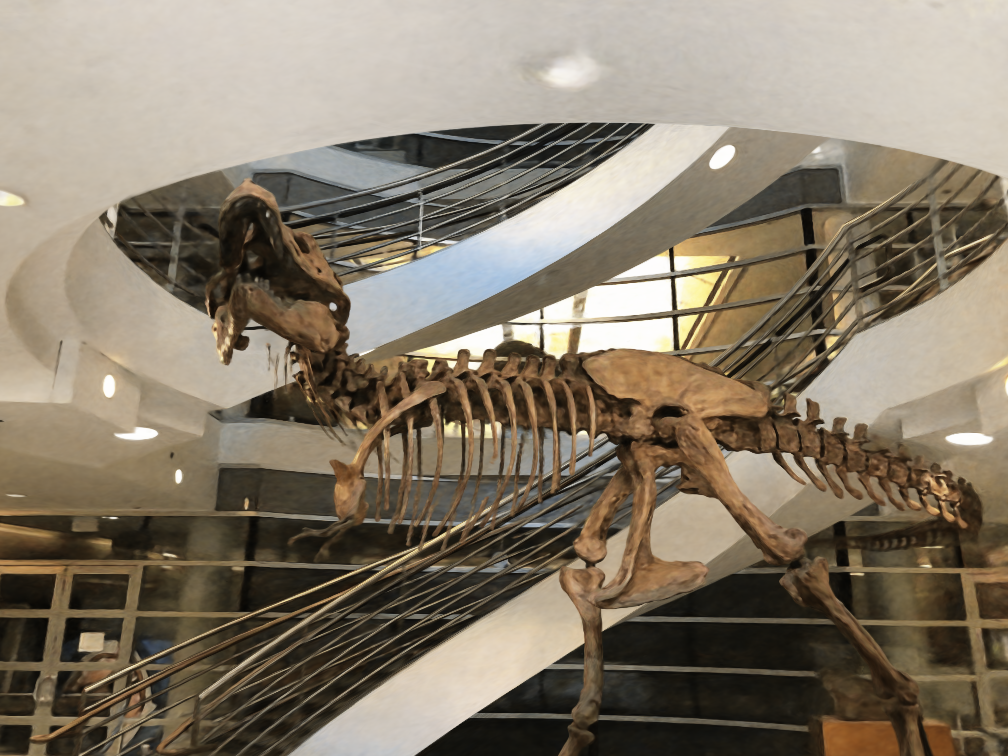}
    \end{minipage}
    \begin{minipage}{.22\textwidth}
        \centering
        \small \textit{w/o SIREN}
        \includegraphics[width=\textwidth]{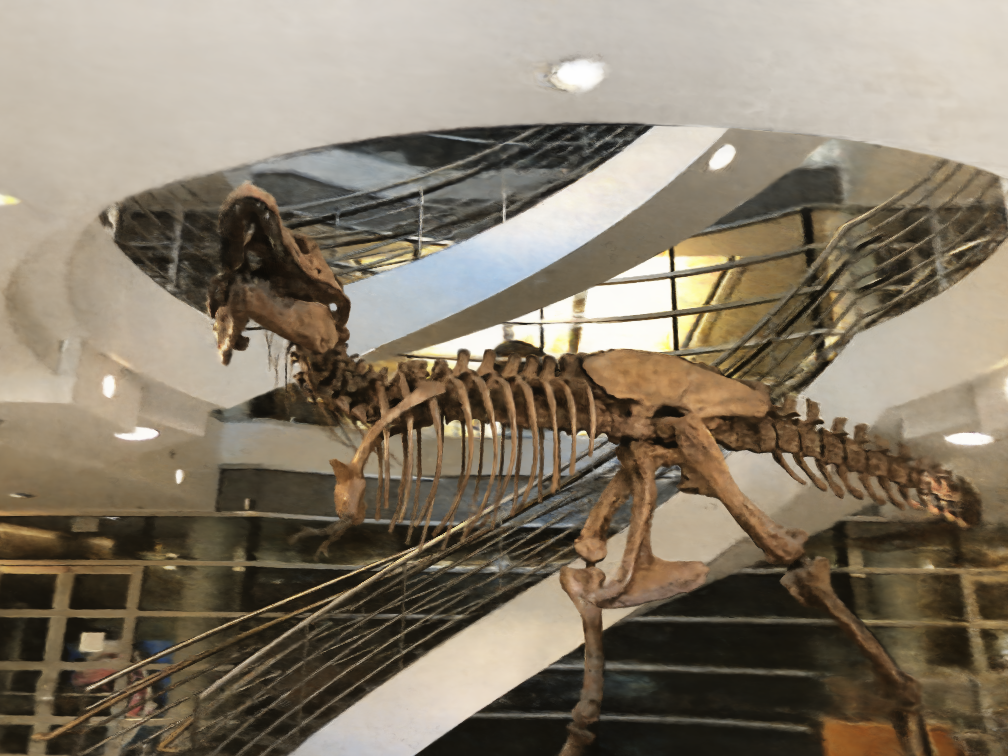}
    \end{minipage}
    \begin{minipage}{.22\textwidth}
        \centering
        \small \textit{w/o MRS}
        \includegraphics[width=\textwidth]{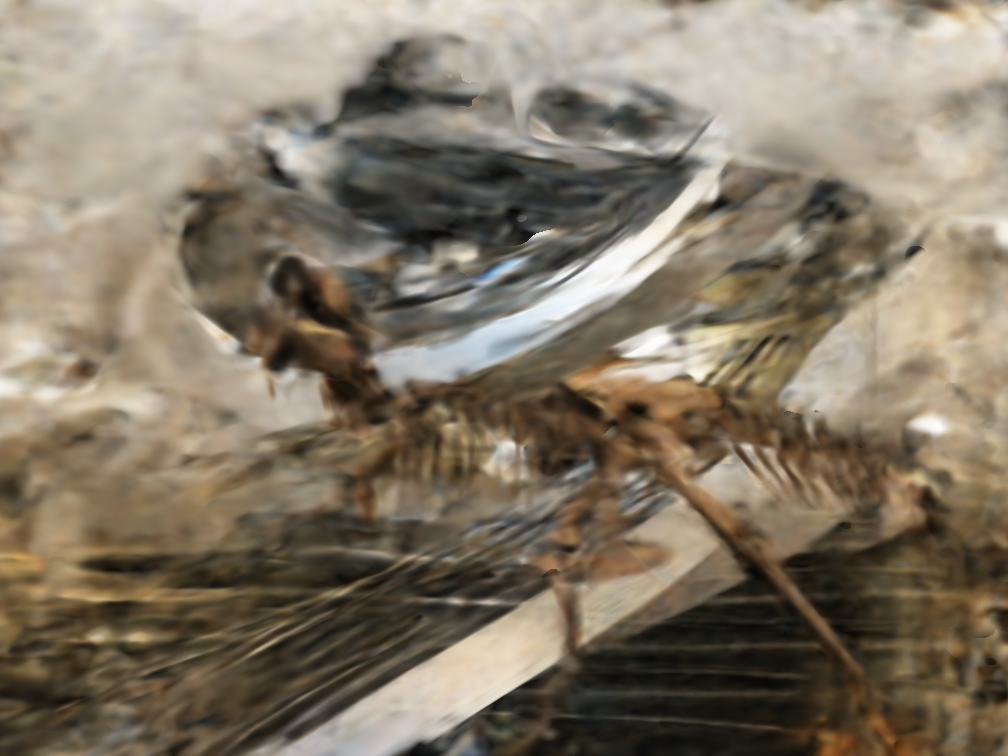}
    \end{minipage}
    \begin{minipage}{.22\textwidth}
        \centering
        \small \textit{w/o SIREN and MRS}
        \includegraphics[width=\textwidth]{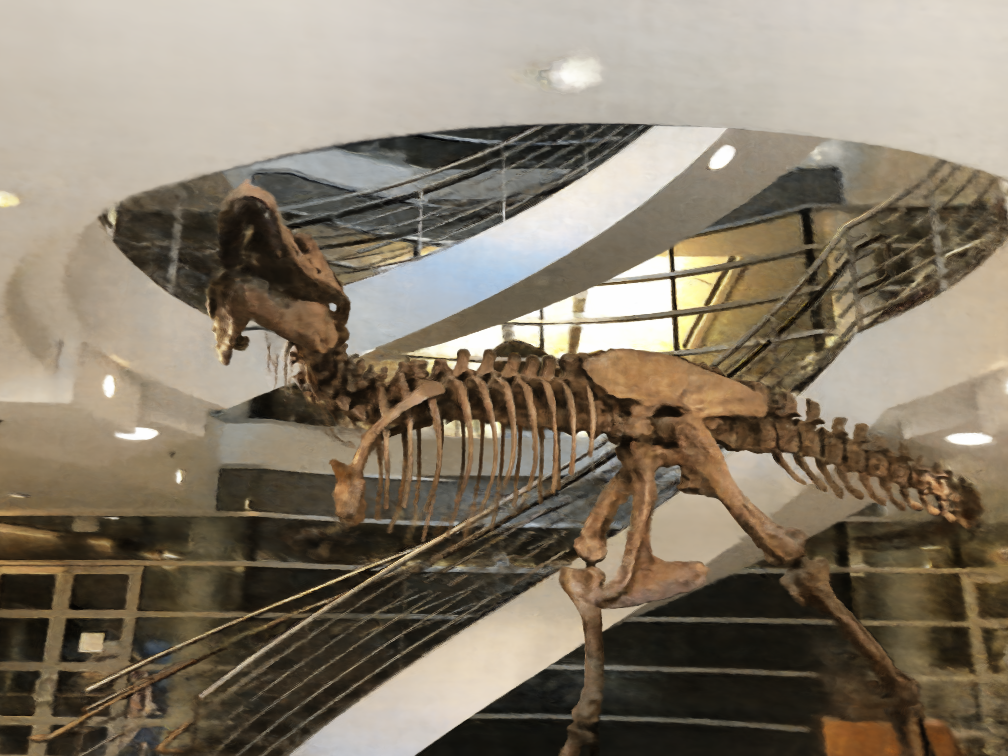}
    \end{minipage}\\
    \vspace{0.05cm} 
    \begin{minipage}{.025\textwidth}
        \rotatebox{90}{Depth}
    \end{minipage}%
    \begin{minipage}{.22\textwidth}
        \centering
        \includegraphics[width=\textwidth]{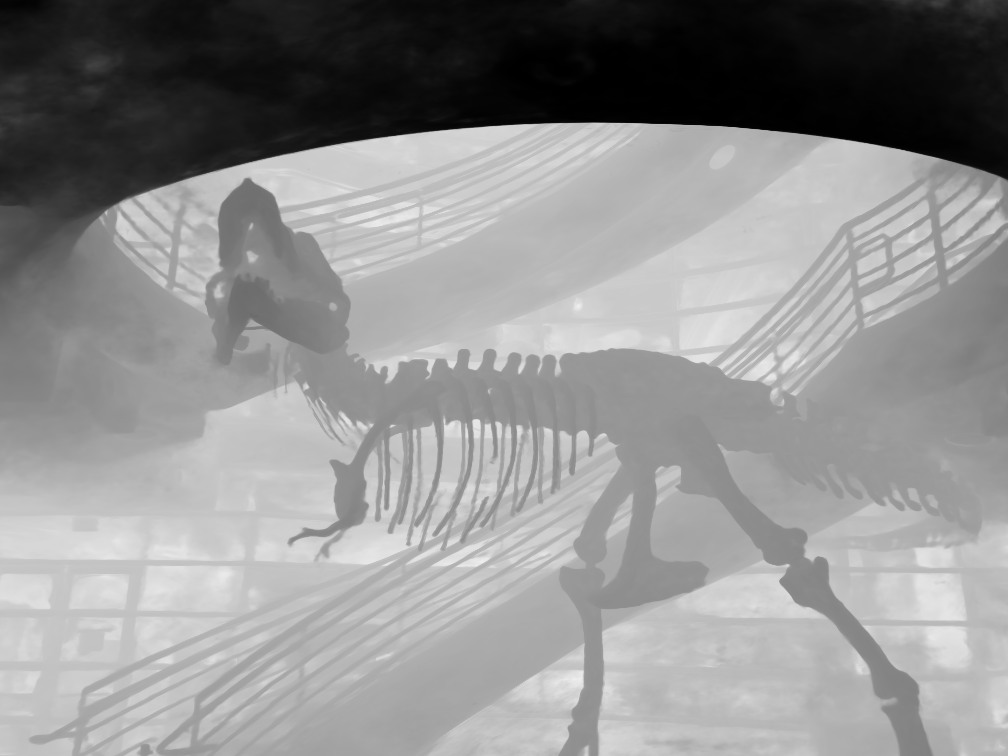}
    \end{minipage}
    \begin{minipage}{.22\textwidth}
        \centering
        \includegraphics[width=\textwidth]{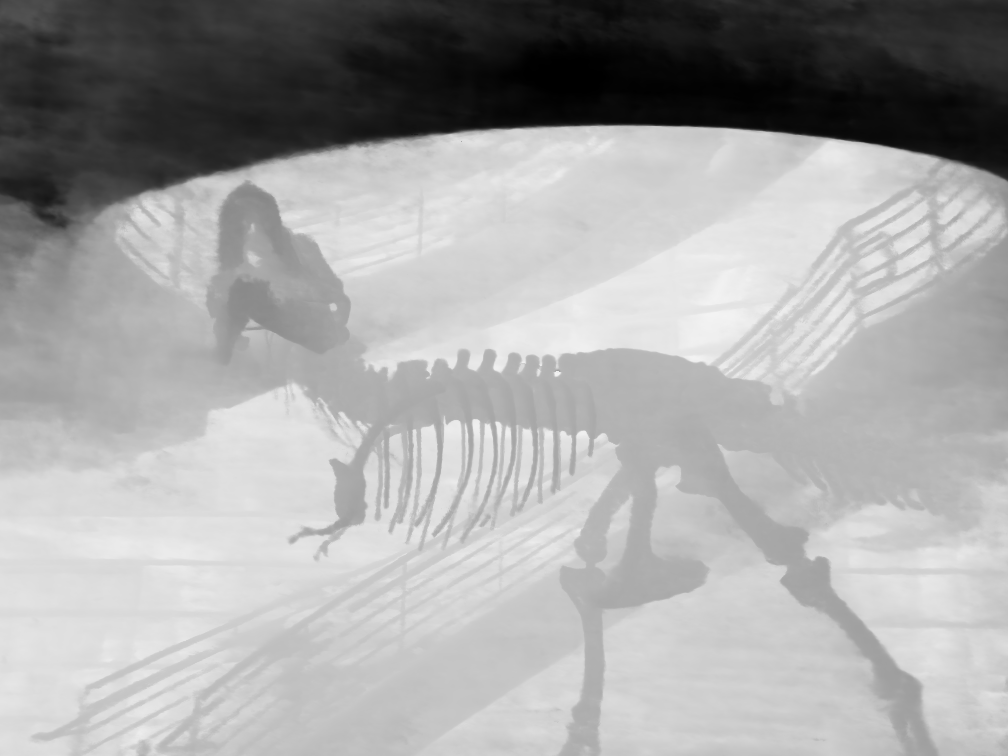}
    \end{minipage}
    \begin{minipage}{.22\textwidth}
        \centering
        \includegraphics[width=\textwidth]{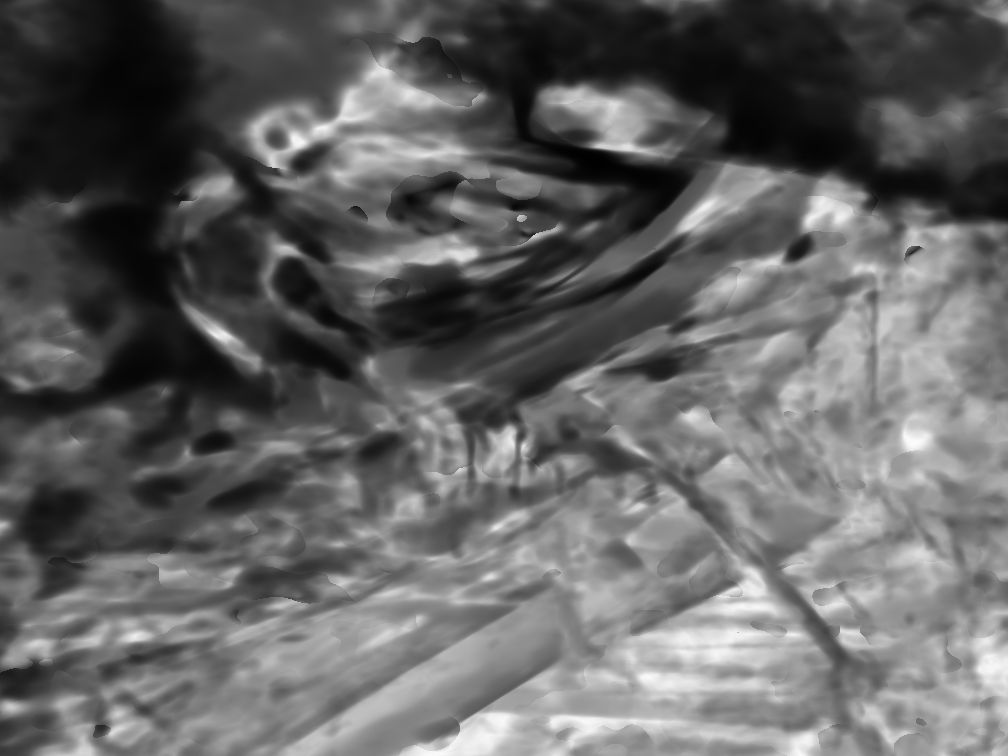}
    \end{minipage}
    \begin{minipage}{.22\textwidth}
        \centering
        \includegraphics[width=\textwidth]{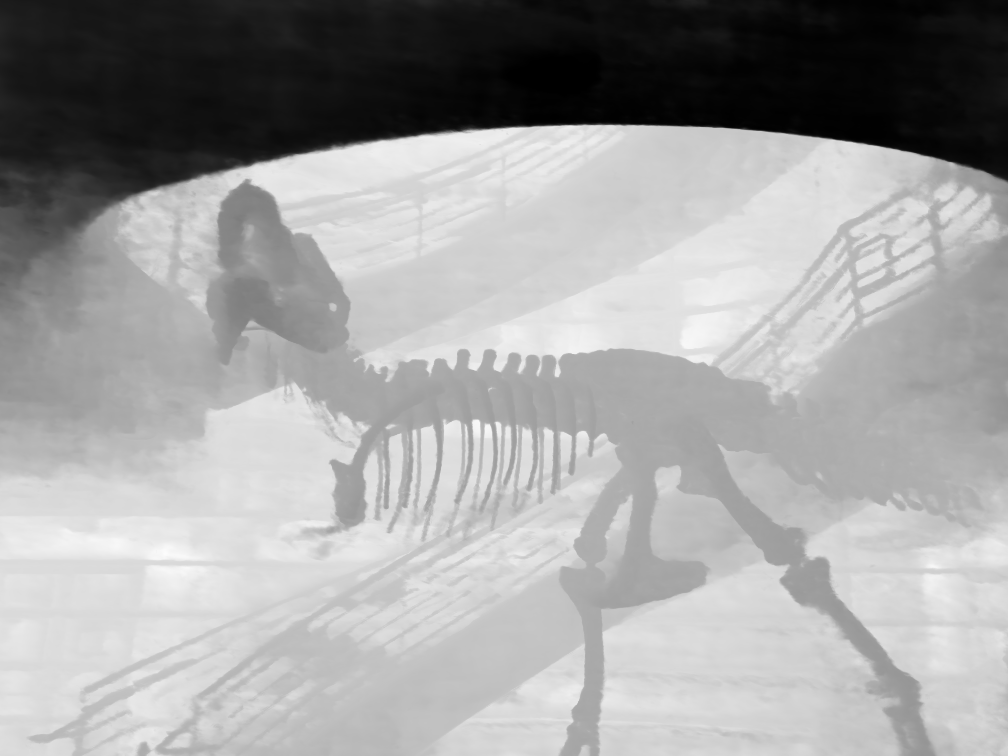}
    \end{minipage}\\
    \begin{minipage}{.025\textwidth}
        \rotatebox{90}{}
    \end{minipage}%
    \begin{minipage}{.22\textwidth}
        \centering
        \small \textbf{PSNR: 24.939}
    \end{minipage}
    \begin{minipage}{.22\textwidth}
        \centering
        \small PSNR: 22.478
    \end{minipage}
    \begin{minipage}{.22\textwidth}
        \centering
        \small PSNR: 14.984 
    \end{minipage}
    \begin{minipage}{.22\textwidth}
        \centering
        \small PSNR: 23.376
    \end{minipage}
    \caption{Qualitative results of ablation study on \textit{Trex} scene.}
    \label{fig_ablation}
\end{figure}

To prove the effectiveness of MRS, we conduct the ablation study on two challenging and representative scenes: $Fortress$, on which NeRFmm is reported to have difficulties converging, and $Trex$, which has lots of fine details and causes relative high pose errors in baselines. Quantitative metrics are shown in Table\,\ref{tab_ablation}. Qualitative results are shown in Figure\,\ref{fig_ablation}.

\noindent\textbf{Analysis on results on \textit{Fortress}.} As mentioned in Section \ref{sec:mrs}, the sampling strategy's diversity is critical for efficient training on textureless scenes like \textit{Fortress}, where Random Sampling may happen to produce a homogenized ray batch that leads to poor supervision. 
MRS, which selects ray batches within a limited amount of key-point regions, would be even more likely to produce undiversified ray batches. Using SIREN-MLP can smoothen the optimization surface and help escape from early local minima. But SIREN-MLP doesn't always guarantee comprehensive improvements.
This explains performances of "SiNeRF" \textgreater "\textit{w/o SIREN and MRS}" or "\textit{w/o MRS}" \textgreater "\textit{w/o SIREN}" in Table \ref{tab_ablation}.

\noindent\textbf{Analysis on results on \textit{Trex}.} SIREN-MLP favors scenes with large consistent patterns like \textit{Flowers} and \textit{Fortress}. Yet on fine-detailed scenes like \textit{Trex}, SIREN-MLP will struggle to distinguish close-by pixels with various colors and produces blurred consistent patterns, as shown in Figure \ref{fig_ablation}. Besides, MRS provides the best performances in combination with SIREN-MLP.
This explains performances of "SiNeRF" \textgreater "\textit{w/o SIREN and MRS}" \textgreater "\textit{w/o MRS}" or "\textit{w/o SIREN}" in Table \ref{tab_ablation}.

\section{Conclusion}
In this work, we identify the potential sources of the systematic sub-optimality of joint optimization. We propose \textit{SiNeRF} architecture and \textit{Mixed Region Sampling} for alleviating such sub-optimality. Experiments and ablation studies show comprehensive improvements in both image synthesis quality and pose accuracy compared to NeRFmm baselines and prove the effectiveness of our designs.

\section*{Acknowledgements}
This work was partly supported by ETH General Fund, the Alexander von Humboldt Foundation and a Huawei research project.

\newpage
\bibliography{egbib}
\end{document}